\pdfoutput=1
\documentclass[11pt]{article}

\usepackage[final]{acl}

\usepackage{times}
\usepackage{latexsym}

\usepackage[T1]{fontenc}
\usepackage[utf8]{inputenc}

\usepackage{microtype}

\usepackage{inconsolata}

\usepackage{graphicx}

\usepackage{balance}
\usepackage{titlesec}
\usepackage{mathrsfs}
\usepackage{amsmath,amssymb,amsfonts,amsthm}
\usepackage{enumitem} 
\usepackage{graphicx}
\usepackage{extarrows}
\usepackage{amssymb}
\usepackage{subfigure}
\usepackage{verbatim}
\usepackage{makecell}
\usepackage{diagbox}
\usepackage{bm}
\usepackage{multirow}
\usepackage{float}
\usepackage{placeins}
\usepackage[linesnumbered, ruled]{algorithm2e}
\SetAlFnt{\small}
\usepackage{epstopdf}
\usepackage{setspace}
\usepackage[super]{nth}
\usepackage{fancyhdr}
\usepackage{lipsum,multicol}
\usepackage{url}
\usepackage{xcolor}
\usepackage{booktabs}
\usepackage{diagbox}
\usepackage[skip=2pt]{caption}
\usepackage{ragged2e}
\usepackage{wrapfig}
\usepackage{tikz}
\usepackage{hyperref}
\usepackage{graphicx}
\usepackage[normalem]{ulem}

\def \eg {\emph{e.g.}, }
\def \ie {\emph{i.e.}, }

\def\BibTeX{{\rm B\kern-.05em{\sc i\kern-.025em b}\kern-.08em
    T\kern-.1667em\lower.7ex\hbox{E}\kern-.125emX}}

\SetCommentSty{mycommfont}

\newcommand{\method}{{Pre$^3$}\xspace}

\allowdisplaybreaks[4]

\title{\method: Enabling Deterministic Pushdown Automata for \\ Faster Structured LLM Generation}
\usepackage[symbol]{footmisc} % 加载宏包，使用符号而非数字
\author{
 \textbf{Junyi Chen\textsuperscript{1}\footnotemark[1]},
 \textbf{Shihao Bai\textsuperscript{2,3}},
 \textbf{Zaijun Wang\textsuperscript{3}},
 \textbf{Siyu Wu\textsuperscript{2}},
 \textbf{Chuheng Du\textsuperscript{1}},
 \\
 \textbf{Hailong Yang\textsuperscript{2}},
 \textbf{Ruihao Gong\textsuperscript{2,3}\footnotemark[2]},
 \textbf{Shengzhong Liu\textsuperscript{1}\footnotemark[2]},
 \textbf{Fan Wu\textsuperscript{1}},
 \textbf{Guihai Chen\textsuperscript{1}}
\\
 \textsuperscript{1}Shanghai Jiao Tong University\hspace{1cm}
 \textsuperscript{2}Beihang University\hspace{1cm}
 \textsuperscript{3}Sensetime Research
\\
 \texttt{\{junyi.chen, dch7723, shengzhong\}@sjtu.edu.cn},
\\
 \texttt{\{wusiyu, hailong.yang, gongruihao\}@buaa.edu.cn},
\\
 \texttt{\{wangzaijun, baishihao\}@sensetime.com}, 
 \texttt{\{fwu,gchen\}@cs.sjtu.edu.cn}
 \\
}

\begin{document}
\maketitle
\footnotetext[1]{Work done during the internship at Sensetime Research.}
\footnotetext[2]{Corresponding authors.}

\begin{abstract}
Extensive LLM applications demand efficient structured generations, particularly for LR(1) grammars, to produce outputs in specified formats (\eg JSON). 
% While existing methods focus on optimizing mask computation, they fail to leverage the holistic structure of the grammar, leading to inefficiencies in rule transitions and context-dependent token processing. 
%Existing methods fail to leverage the holistic grammar structure, leading to runtime overhead in rule transitions and context-dependent token processing, which is especially inefficient under large inference batches. 
Existing methods primarily parse LR(1) grammars into a pushdown automaton (PDA), leading to runtime execution overhead for context-dependent token processing, especially inefficient under large inference batches.
% Moreover, large inference batches lead to more transitions and context-dependent token computation, causing excessive runtime overhead.
To address these issues, we propose \method that exploits deterministic pushdown automata (DPDA) to optimize the constrained LLM decoding efficiency. 
%First, \method constructs a unified pushdown automaton as grammar representations, replacing complex reductions with lightweight edge transitions with minimal overhead. 
First, by \textbf{pre}computing \textbf{pre}fix-conditioned edges during the \textbf{pre}processing, \method enables ahead-of-time edge analysis and thus makes parallel transition processing possible. 
%Second, by preprocessing execution stack conditions into transition edges offline, \method eliminates runtime path exploration and enables edge optimization and parallel transition processing. 
Second, by leveraging the prefix-conditioned edges, \method introduces a novel approach that transforms LR(1) transition graphs into DPDA, eliminating the need for runtime path exploration and achieving edge transitions with minimal overhead.
\method can be seamlessly integrated into standard LLM inference frameworks, reducing time per output token (TPOT) by up to 40\% and increasing throughput by up to 36\%  in our experiments. Our code is available at \url{https://github.com/ModelTC/lightllm}.
\end{abstract}
\section{Introduction}
%\red{nubmers of structured generation APIs}
The recent remarkable development of Large Language Models (LLM) has ushered in new opportunities for a wide array of intelligent applications such as automated external tool invocations through function calls~\cite{cai2023large, li2024large, zhuo2024bigcodebench}, chain of thoughts~\cite{wei2022chain, wang2022self, openaio1, guo2025deepseek}, embodied AI~\cite{duan2022survey, brohan2023rt, yang2024embodied} et al. These applications created substantial demand for LLM systems to perform structured generation and produce outputs adhering to specific formats, such as JSON or other structures. Notably, major LLM API providers such as OpenAI and Alibaba Cloud now support JSON mode output to ensure deterministic schema compliance. Downstream applications can accordingly utilize these structured outputs to engage in downstream system interactions~\cite{cho-etal-2023-discrete}.
% \highlight{add references}

% 这一段话的逻辑：约束解码常见->约束解码复杂->约束解码引入了很大开销
% Constrained decoding is a common technique in structured generation, where invalid tokens are excluded by applying a \textit{probability mask}—setting their probabilities to zero.To accommodate diverse structural formats across applications, a flexible mechanism is needed to specify and verify constraints. LR(1) grammars~\cite{francis1961qr} offer a general approach by defining structures through recursive rules, enabling complex combinations that surpass the flexibility of regular expressions. However, this very flexibility makes direct application to constrained decoding inefficient. Each decoding step involves parsing the grammar for every possible token in a potentially large vocabulary, which can lead to significant overhead. Furthermore, since LLM-generated tokens are derived from the vocabulary and may consist of multiple characters that can extend beyond grammar rule boundaries, the execution stack must specifically manage this by carefully handling accept and reduction operations.

\textit{Constrained decoding}~\cite{hu-etal-2019-improved, scholak2021picard} is a widely used method in structured generation tasks~\cite{willard2023efficientguidedgenerationlarge, dong2023codep, ruckstiess2024origami} that excludes invalid tokens at each step by applying a \textit{probability mask} to zero out their sample possibility. Flexible mechanisms like LR(1) grammars~\cite{francis1961qr, KNUTH1965607} are often employed to handle diverse and complex structural constraints, as they allow recursive rule definitions that surpass the limitations of regular expressions. However, this flexibility comes at the cost of degraded efficiency: Each decoding step requires parsing the grammar for all candidate tokens in a potentially large vocabulary. Additionally, tokens generated by LLM may consist of multiple characters that span across grammar rule boundaries, further complicating the generation process and demanding dedicated execution stack management. Both of them lead to significant computational overhead. These challenges raise the need to optimize constrained decoding efficiency without affecting LLM generation fidelity, making it more applicable in real-world applications.

Current state-of-the-art (SOTA) methods for constrained decoding acceleration, such as XGrammar~\cite{dong2024xgrammarflexibleefficientstructured}, primarily focus on parsing LR(1) grammars into a pushdown automaton (PDA)~\cite{nederhof1996efficient}. A PDA consists of multiple finite state automata (FSA), each representing a grammar rule, with the stack handling recursive rule expansions. These methods achieve substantial speedups by precomputing masks while managing transitions through pushdown automata. However, they overlook the inherent properties of LR(1) grammars, which can be equivalently transformed into a deterministic pushdown automaton (DPDA)~\cite{valiant1973decision, valiant1975regularity}.
%By parsing LR(1) grammars into a DPDA, rather than focusing solely on optimizing mask computations within the PDA framework, we facilitate more efficient and scalable constrained LLM decoding.

% v2
%Traditional PDA-based approaches~\cite{} compute probability masks and transitions independently for grammar rules using separate pushdown automata. However, this design fails to leverage the holistic grammar structure, resulting in two key limitations.
%\highlight{Better to define ``holistic grammar structure'' and distinguish it from separate grammar rules.}
%First, treating rules in isolation introduces unnecessary complexity in the transition steps between rules, requiring meticulous stack management and causing excessive overhead. 
%Second, the precomputed masks are incomplete for certain tokens, called \textit{context-dependent tokens}, whose mask computation relies on runtime information that is only available during inference. 
%As inference batch size grows, both the number of transitions and context-dependent tokens with runtime computation increase significantly, introducing overhead that existing techniques cannot fully address, making overall decoding efficiency severely degraded.

The primary issue with traditional PDA-based approaches~\cite{koo2024automata, park2025flexible, dong2022codepad, willard2023efficient, li2024formal} stems from the non-deterministic nature of the PDA's edges. Although these methods precompute masks based on the PDA structure, this design introduces two critical limitations.
First, the non-deterministic edges depend on runtime contextual information to resolve transitions, resulting in incomplete precomputed masks for \textit{context-dependent tokens}. The computation of context-dependent tokens necessitates backtracking, speculative operations, and the maintenance of a \textit{persistent stack} (merges all past stacks into a tree, with each stack as a root-to-node path) during runtime. As batch sizes increase, the overhead from these runtime computations grows significantly, severely degrading decoding efficiency.
Second, previous methods cannot effectively optimize non-deterministic transitions during preprocessing because they will dynamically change during runtime. This limitation hinders their ability to fully exploit the potential of the parsing method, leading to suboptimal performance. % By addressing these issues, we aim to enable more efficient and scalable constrained decoding.

To address these challenges, we propose \method, a constrained LLM decoding approach based on a deterministic pushdown automaton (DPDA). Unlike traditional methods, we design an algorithm to directly build a DPDA from the LR(1) grammar. Leveraging the deterministic nature of the DPDA's edges, our approach resolves the aforementioned limitations.
First, the determined transitions in the DPDA eliminate the context-dependent tokens, further entirely eliminating the need for backtracking, speculative exploration, and the maintenance of a persistent stack. This fundamentally reduces the runtime computational overhead associated with transitions.
Second, since all transition edges in the DPDA are available during preprocessing, we can perform comprehensive optimizations on the automaton in advance. Additionally, for the stack-matched transition mechanism of the DPDA, we design a parallel verification method for transitions, which accelerates inference.
Together, these innovations result in a more efficient and scalable constrained decoding framework.
% \highlight{This paragraph needs to be further calibrated to highlight and explain the novelty and rationale of your algorithm design in more straightforward manner.}

In summary, the paper's main contributions are:

\begin{itemize}[noitemsep,topsep=1pt,leftmargin=*]
%    \item We propose a method to generate state transition graphs from LR(1) grammars and construct deterministic pushdown automata, eliminating the need for precomputing context-dependent and context-independent masks.
%    \item We enhance the pushdown automata by attaching stack operations to edges and introducing a cycle-handling strategy to prevent infinite reduction loops, enabling direct application to grammar parsing.
%    \item We leverage GPU parallelism to accelerate mask computation for large vocabularies and optimize the transition graph by merging redundant edges, reducing complexity while preserving correctness for unambiguous LR(1) grammars.
    % \item We build a unified pushdown automaton for the entire grammar, turning complex reduction operations into simple state transitions and improving the performance.
    % \item By adding stack matching conditions to transition edges, we enable precomputation of all feasible transitions and make the automaton deterministic, ensuring efficient processing even for large batch sizes.
    
    %\item We build a unified pushdown automaton for the grammar, transforming reduction operations into lightweight edge transitions.
    %\item We introduce stack-aware precomputed transitions, eliminating runtime exploration, enabling efficient and deterministic processing for large batches, and performing edge optimization during preprocessing.
    \item We first propose an algorithm to transform LR(1) state transition graphs into DPDA, eliminating runtime exploration and enabling edge transitions with minimal overhead.
    \item We enable additional optimizations for edges and support parallel transition processing by precomputing prefix-conditioned edges.
    \item We integrate \method into mainstream LLM inference systems and achieve up to 40\% improvement in time per output token (TPOT) and increase throughput by up to 36\% with high scalability into large batch sizes.
\end{itemize}

% For reproducibility, our code is made available on Github\footnotemark[1] to encourage future research in the field.

% \footnotetext[1]{\url{URL Here}}
\section{Preliminaries and Background}

\subsection{LLM Constrained Decoding}

Constrained decoding~\cite{hu-etal-2019-improved} enforces strict grammatical adherence by dynamically pruning invalid tokens during generation. While effective for structural compliance, existing methods face two key challenges: (1) handling diverse grammars, large vocabularies, and complex token-to-text mappings, and (2) computational inefficiency at scale, especially under large batch processing where dynamic validation creates sequential bottlenecks.

Our empirical analysis reveals this critical limitation. When evaluating XGrammar on Meta-Llama-3-8B (2×H800 GPUs), constrained decoding exhibits up to 37.5\% higher latency (147.64 ms vs. 92.23 ms) at batch size 512 compared to unconstrained decoding (Table~\ref{tb:xgrammar-vs-baseline}). The performance gap grows with batch size due to non-parallelizable validation steps, which is a fundamental constraint in current approaches.
\begin{table}[]
\caption{Per token latency comparison (in milliseconds) across different batch sizes.}
\label{tb:xgrammar-vs-baseline}
\resizebox{\linewidth}{!}{
\begin{tabular}{@{}c|cccccc@{}}
\toprule
\textbf{Batch Size} & \textbf{16} & \textbf{32} & \textbf{64} & \textbf{128} & \textbf{256} & \textbf{512} \\ \midrule
Baseline            & 11.38       & 21.87       & 25.74       & 30.08        & 56.29        & 92.23        \\
XGrammar            & 15.19       & 43.69       & 52.07       & 65.21        & 90.98        & 147.64       \\ \bottomrule
\end{tabular}
}
\end{table}
These findings motivate the need for a new constrained decoding approach that maintains grammatical correctness while achieving better computational efficiency at scale. 

\subsection{LR(1) Grammar and State Transition Graphs}
% \highlight{Make this part a little shorter and relate them with LLM}
In constrained decoding scenarios, most grammars can be classified as LR(1) grammars, which are fundamental to bottom-up parsing and align naturally with the token-by-token generation process of large language models (LLMs). LR(1) grammars are a powerful subset of context-free grammars capable of describing the syntax of most programming languages. They are characterized by their ability to handle deterministic parsing with a single lookahead symbol, making them highly expressive and widely applicable. Nearly all context-free grammars can be converted into LR(1) form, which ensures their versatility in modeling structured languages. This property, combined with their alignment with bottom-up parsing methods, makes LR(1) grammars a cornerstone in constrained decoding and syntactic analysis tasks.
%Moreover, LR(1) grammars can be transformed into deterministic pushdown automata, eliminating the distinction between context-dependent and context-independent tokens.

LR(1) items are tuples of the form \([A \to \alpha \cdot B\beta, a]\), where \(A \to \alpha \cdot B\beta\) represents the parsing progress of a production rule, and \(a\) is a lookahead symbol used to determine when a reduction should occur. The \texttt{CLOSURE} operation constructs LR(1) item sets by adding items for non-terminals and their productions, ensuring all possible derivations are considered. The \texttt{GOTO} function generates the LR(1) state transition graph by moving the dot in items past a grammar symbol \(X\) and computing the closure of the resulting items, thereby connecting states to form the LR(1) automata. This process continues until no new states are generated, creating a complete parsing structure for the grammar.

\subsection{Deterministic Pushdown Automata (DPDA)}
Pushdown automata (PDA) are a class of abstract machines that extend finite automata with an unbounded stack memory, enabling them to recognize context-free languages (CFLs)~\cite{hopcroft2001introduction}. A PDA is defined as a 7-tuple \((Q, \Sigma, \Gamma, \delta, q_0, Z_0, F)\), where \(Q\) is a finite set of states, \(\Sigma\) is the input alphabet, \(\Gamma\) is the stack alphabet, \(\delta: Q \times (\Sigma \cup \{\epsilon\}) \times \Gamma \rightarrow \mathcal{P}(Q \times \Gamma^*)\) is the transition function, \(q_0\) is the initial state, \(Z_0\) is the initial stack symbol, and \(F \subseteq Q\) is the set of accepting states. The non-deterministic transition function \(\delta\) allows PDAs to handle ambiguous structures inherent to context-free grammars, \eg nested parentheses or recursive syntactic patterns. 

A deterministic pushdown automaton (DPDA) is a restricted variant where, for every state \(q \in Q\), input symbol \(a \in \Sigma\), and stack symbol \(Z \in \Gamma\), the transition function \(\delta(q, a, Z)\) yields at most one possible move, and \(\epsilon\)-transitions (stack operations without consuming input) are permitted only if no input-consuming transition is available~\cite{sipser1996introduction}. This determinism ensures unique computation paths, making DPDAs equivalent to the class of deterministic context-free languages (DCFLs), which are unambiguous and efficiently parsable. As mentioned earlier, the vast majority of grammars in the constrained decoding scenario can be represented by LR(1), which is a true subset of DCFL and can be recognized by DPDA~\cite{asu861986compilers, sipser1996introduction}. Compared to PDA, DPDA avoided backtracking and non-deterministic search overhead, which can significantly improve the efficiency of constrained decoding. See Appendix~\ref{sec:appendix-1} for additional background on formal language theory.

% 3.1 Prefix-conditioned Edge
% 承上启下过渡段
% 现有内容

% 3.2 DPDA Construction
% 承上启下过渡段：with prefix-conditioned edge, we can xxxxx, to construct xxx. We categorize edge into two types: xxx, and xxx. By adding these types of edges, a DPDA is constructed.

% \textbf{ Definition of acceptance of reduction edge.} 结合图讲清楚
% \textbf{ Adding acceptance edge.} 
% \textbf{ Adding reduction edge.} 
% \textbf{ Deleting Edge.} 
% ref{algorithm}

% 3.3 Optimization on DPDA
% 承上启下过渡段
% Edge Optimization. Aggregation + Merging
% Computation Optimization. Parallel 
\section{\method Design}
\begin{figure*}[t!]
    \centering
    \includegraphics[width=0.8\linewidth]{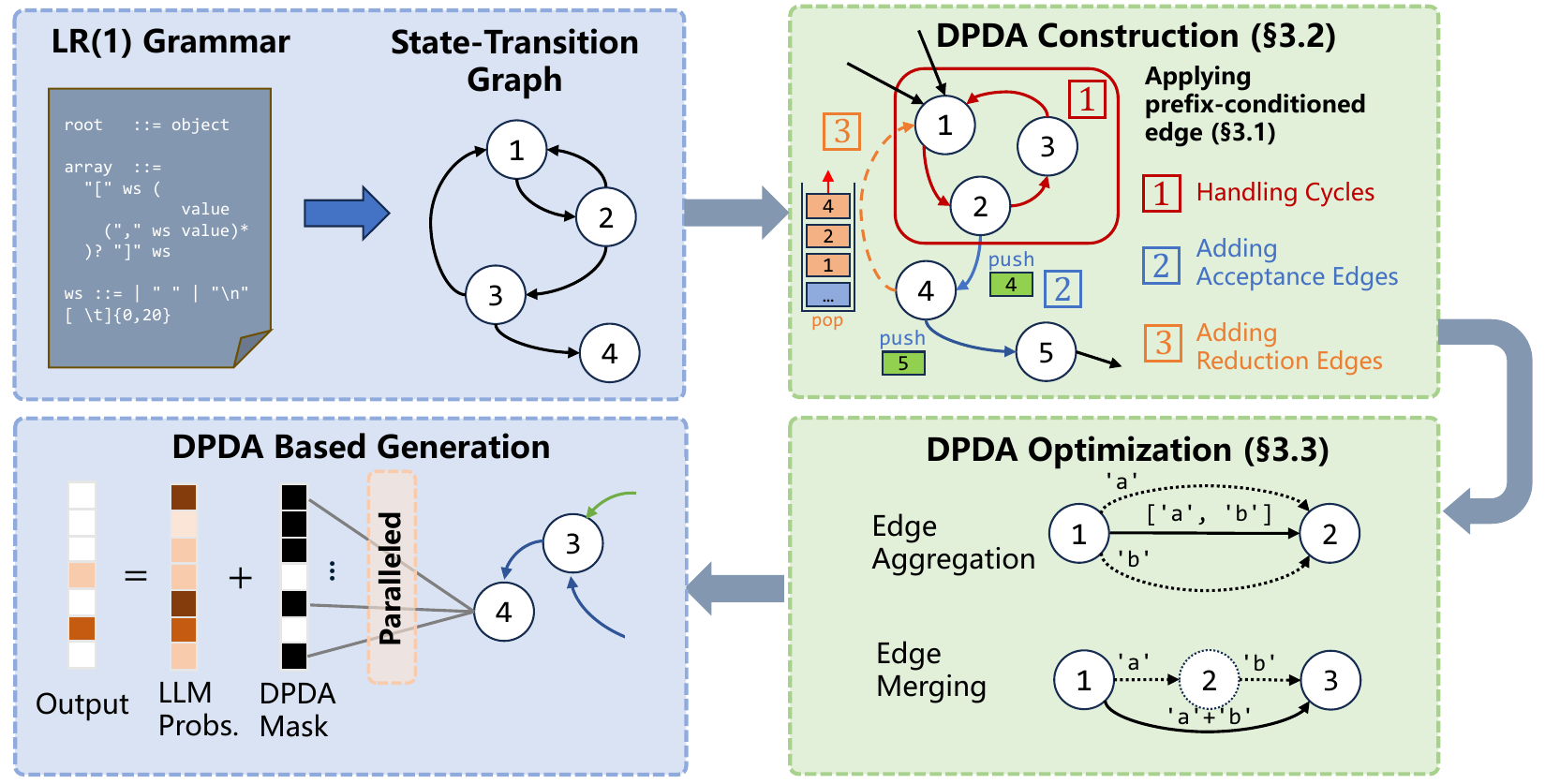}
    \caption{Overview of \method: The figure depicts the workflow from LR(1) grammar to DPDA-based generation, encompassing DPDA construction and optimization steps.}
    \label{fig:method_overview}
\end{figure*}
% 我们提出的\method是一个基于状态机转移的约束解码方案，也可以称作graph-based解法。我们的方法通过将LR(1)文法转换成LR(1)状态转移图，然后再利用3.2和3.3小节中的方案将其转化为一个可以直接用于约束解码的确定性下推自动机。完整的处理流程如\ref{fig:method_overview}所示
% Constrained decoding with large language models (LLM) presents significant challenges, particularly when enforcing strict adherence to complex grammars. Traditional approaches, such as mask-based methods, often struggle with efficiency and scalability due to the need for runtime stack inspection and the handling of context-dependent tokens. These limitations become especially pronounced in large-batch scenarios, where computational overhead can severely degrade performance. To address these issues, we propose a novel framework that leverages deterministic pushdown automata (DPDA) to enable efficient and scalable constrained decoding.
Our proposed method, \method, is a DPDA-based constrained decoding solution that leverages a novel approach for constructing a DPDA from a given LR(1) grammar. The method operates by first transforming the LR(1) grammar into an LR(1) state transition graph, which is then converted into a DPDA using the techniques introduced in this section. This DPDA can be directly utilized for constrained decoding, enabling efficient and effective decoding. The method requires only minimal processing time, averaging 3-5 seconds for complex JSON grammars and under 0.1 seconds for simpler grammars (\eg arithmetic expressions). Notably, this is a one-time cost as the results are cacheable and reusable. The complete workflow of our method is illustrated in Figure \ref{fig:method_overview}.

In Section~\ref{section-3-1}, we introduce the Prefix-conditioned Edge, a novel mechanism ensuring uniqueness by matching both prefix information and input symbols, unlike traditional PDA transitions. In Section~\ref{section-3-2}, we design an algorithm to compute all LR(1) state transitions, incorporating Prefix-conditioned Edge and addressing cyclic structures, successfully constructing a DPDA. In Section~\ref{section-3-3}, we optimize the DPDA's structure and performance through preprocessing, leveraging its pre-determined edges.

\subsection{Prefix-conditioned Edges} \label{section-3-1}
Constrained decoding with LLMs faces challenges due to non-deterministic transitions in PDA, where the same input symbol can trigger multiple transitions based on prior symbol sequences. This non-determinism complicates computation by requiring speculative exploration, backtracking, and a persistent stack to store historical context, increasing overhead. To resolve these issues, eliminating non-determinism in transitions is crucial for enabling preprocessing optimizations and efficient runtime computation.
%During preprocessing, the inability to determine the target of a transition edge prevents optimizations such as aggregation or merging, limiting the efficiency of the automaton's structure. 

%Recall that LR(1) grammars have a useful property: \textbf{by examining the current stack and looking ahead just one symbol, we can uniquely determine whether to shift or reduce and which production rule to apply.}

\begin{figure}[t!]
    \centering
    \includegraphics[width=0.85\linewidth]{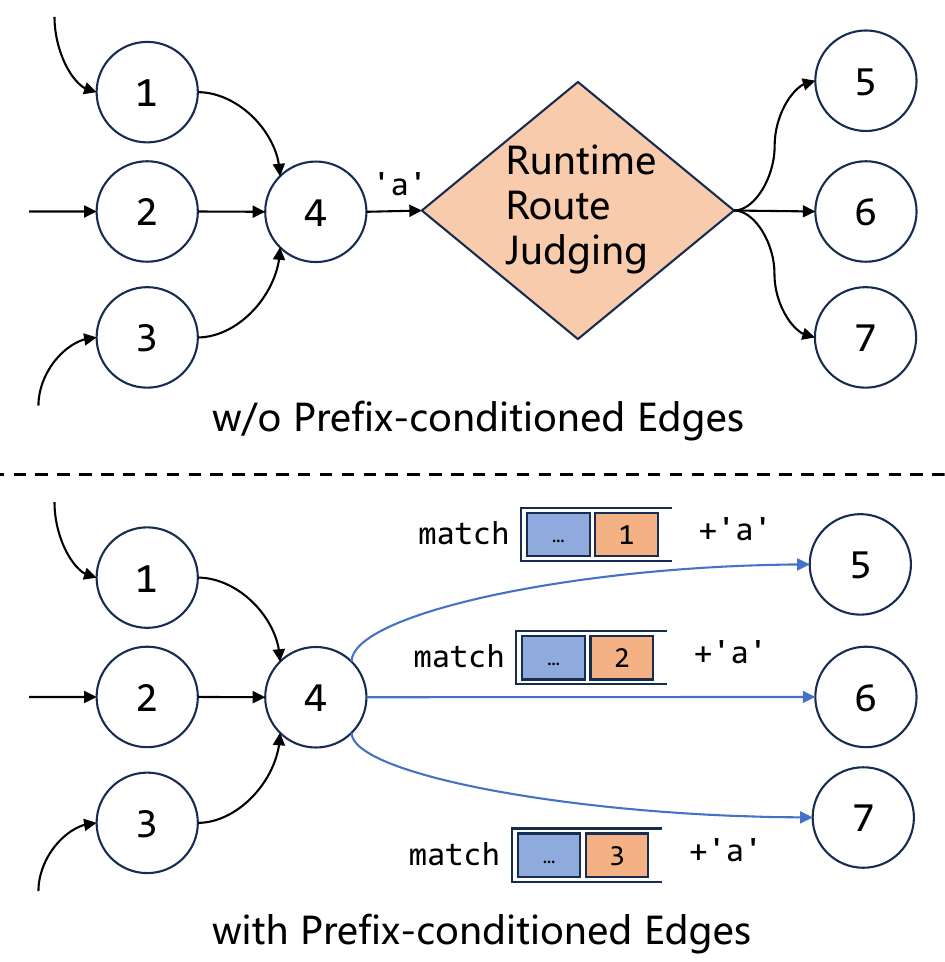}
    \caption{This diagram illustrates prefix-conditioned edges: above shows the case before calculation, where `a' is a context-dependent token requiring runtime context for transition; below shows the precomputed case, where each edge includes a stack-matching condition, uniquely determining the transition path via the condition and transition symbol.}
    \label{fig:prefix-conditioned-edge-description}
\end{figure}

A fundamental property of LR(1) grammars is that \textbf{the current stack configuration and a single lookahead symbol are sufficient to uniquely determine the next action}. This property provides a theoretical foundation for introducing determinism into the automaton's transition edges. Building on this insight, we propose the Prefix-conditioned Edge, as illustrated in Figure~\ref{fig:prefix-conditioned-edge-description}. 

By simultaneously considering the input symbol and the prefix of accepted symbols (represented by the stack's state), we uniquely determine the target state for each transition. To achieve this, our method enhances each edge with three key components:

\begin{itemize}[noitemsep,topsep=1pt,leftmargin=*]
\item \textbf{Accepted Symbol}: The input symbol that triggers the transition.
\item \textbf{Stack Matching Condition}: The specific prefix of the stack required for the transition to be valid.
\item \textbf{Stack Operations}: Actions such as push to update the stack during the transition, which is both required by PDAs and DPDAs.
\end{itemize}

% 在LLM约束解码中，之所以一些context-dependent token的部分mask计算需要在运行时根据执行栈的上下文实时处理，这主要是因为LR(1)文法在仅根据下一个输入的符号时，部分情况下无法确定下一步的执行动作
%Building on this insight, we introduce Prefix-conditioned edge, illustrated in Figure~\ref{fig:prefix-conditioned-edge-description}. This approach integrates the prefix of accepted symbols (represented by the stack's state) as transition conditions into the automaton's edges. Specifically, for each possible transition, the method uniquely determines the target state based on the current stack configuration and input symbol. By explicitly encoding these conditions into the edges, the automaton gains determinism—each transition is precisely defined without ambiguity. This eliminates the computational overhead discussed earlier.

% v2
%By embedding these components, we precompute all possible transitions along with their corresponding stack conditions. This precomputation ensures that the automaton operates deterministically, eliminating the need for runtime backtracking. Context-dependent tokens are resolved statically during the precomputation phase, resulting in highly efficient and predictable decoding.

Notably, although the additional stack-matching conditions introduced to the edges increase complexity, we address this challenge by implementing a parallel algorithm capable of simultaneously verifying multiple stack-matching conditions, effectively resolving this issue.

%During mask precomputation, we know the next symbol due to the fixed LLM vocabulary. However, the stack in the model typically stores the state nodes it has passed through, meaning that the current state of the stack is not accessible during precomputing.
%Knuth proposed that the language of LR(k) grammars is exactly a subset of the languages accepted by Deterministic Pushdown Automata (DPDA), so LR(1) grammar can always be transformed into an equivalent DPDA. 
%This means that, although the calculation of some tokens needs to be processed based on the information of the execution stack, the paths of their transitions are limited and determined. That is, we can precompute all possible situations of the context stack during the transition according to LR(1).
% 那么这就意味着，虽然一些token的计算需要根据执行栈的信息进行处理，但是他们转移的路径都是确定的有限个，即我们可以根据LR(1)预先计算出转移时上下文栈的所有可能情况

%Based on the analysis above, we propose Prefix-conditioned Edge Precomputation where the contents of the stack are incorporated into the transition conditions of the automaton’s edges. This allows us to precompute all possible transitions ahead of time, removing the need to dynamically inspect the stack. In doing so, context-dependent tokens no longer arise, as all decisions are fully determined by the precomputed stack conditions and the lookahead symbol. Once the automaton become deterministic, the context-dependent tokens will disappear.

% \subsection{Transform the LR(1) State Transition Graph into a Deterministic Pushdown Automaton}
% \subsection{Precomputation of Prefix-conditioned Transition Edge} \label{section-3-2}
%The LR(1) state transition graph alone is insufficient for efficient state transitions due to the presence of both terminal and non-terminal edges. Handling non-terminal transitions requires scanning item sets, selecting reduction rules, backtracking along the graph, and then proceeding with the transition, introducing significant computational overhead, especially for complex grammars with large item sets and long production rules.
% 从理论上，Knuth提出LR(k)文法的语言也恰好是DPDA接受语言的一个子集。因此LR(1)文法必定可以被转化成一个与其等价的确定性下推自动机。回想context-dependent token的来源在于，precompute mask的过程中，LLM的词表固定，因此我们可以知道下一个符号是什么，而运行栈中通常保存着当前推理经过的状态节点，因此此时我们无从得知当前运行栈的信息。LR(1)文法具有一个非常优秀的性质，只需要根据当前运行栈中的内容并额外向后查看输入串的1个符号，就可以唯一确定动作是移进还是归约、利用相应产生式进行归约。那么我们提出可以将当前运行栈中的内容作为转移边的转移条件的一部分，这样就可以提前预先计算出所有可行的转移边。利用这种方法，Context-dependent类型的token就不再存在了

\subsection{Cycle-aware Deterministic Pushdown Automata Construction} \label{section-3-2}
% In theroy, it has also been proved~\cite{} that LR(k) grammars, including LR(1), can be equivalently transformed into deterministic pushdown automata (DPDA), allowing for fully deterministic parsing. Since traditional PDA-based method suffered from the non-deterministic property of the transition edge, in this section we will introduced our proposed DPDA construction algorithm by applying the prefix-conditioned to LR(1) state transition graph.
To avoid the additional exploration overhead at runtime, we aim to construct a DPDA based on LR(1) grammars. However, building a DPDA is non-trivial and requires a systematic approach. In this section, we introduce our algorithm for constructing a DPDA from an LR(1) state transition graph step by step, leveraging the prefix-conditioned edge to ensure determinism.

\subsubsection{DPDA Structure} 
\begin{figure}[t!]
    \centering
    \includegraphics[width=0.75\linewidth]{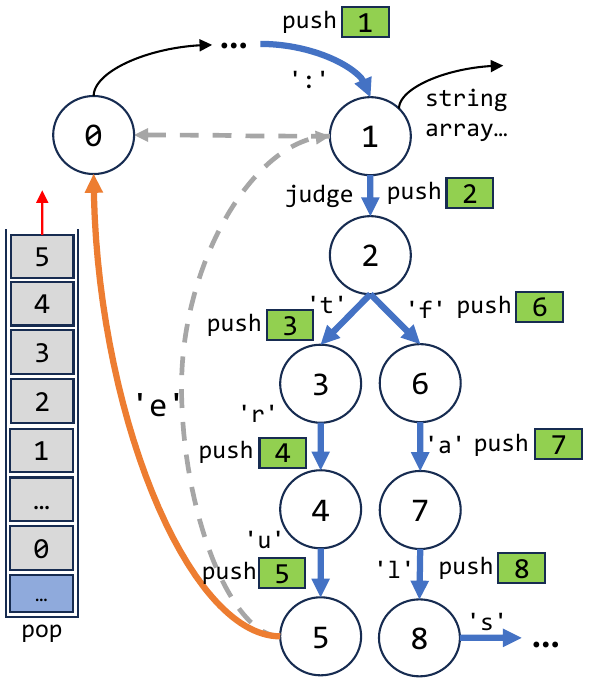}
    \caption{Two edge types for DPDA computation: blue edges are acceptance edges (existing in the original LR(1) graph, handling stack operations for acceptance); orange edges are reduction edges (added to the DPDA, matching and popping stack operations for reductions); gray edges depict LR(1) reduction paths, demonstrating fewer nodes needed for reduction after state machine construction.}
    \label{fig:automaton_construction}
\end{figure}

%Traditional separated pushdown automata (PDA) face limitations in handling transitions across multiple grammar rules. While the LR(1) state transition graph contains all grammatical information, it fails to fully address all transition paths, making it insufficient for efficient inference. 
%To address the limitations, we construct a unified LR(1) grammar-based deterministic pushdown automaton (DPDA). The nodes of this DPDA are derived from the LR(1) item set family in the state transition graph. To build the automaton, we compute all possible transitions between nodes and distinguish two edge types: \textit{acceptance edges} and \textit{reduction edges}, and we further introduce prefix-conditioned edges (as defined in Section~\ref{section-3-2}) to ensure the deterministic.
We begin our algorithm with the state transition graph generated from the LR(1) grammar, where the nodes represent the LR(1) item set family and the edges indicate the acceptance of a symbol when traversing from one node to another. Building on this foundation, we construct the DPDA by retaining the node definitions from the LR(1) transition graph but redefining the edges into two distinct types: \textit{acceptance edges} and \textit{reduction edges}, as shown in Figure~\ref{fig:automaton_construction}.

\begin{itemize}[topsep=0pt,leftmargin=0.35cm]
    \item \textbf{Acceptance Edges} are the simplest type of transition in our DPDA. These edges are directly derived from the original state transition graph of the LR(1) grammar. In the context of LR(1) parsing, an acceptance edge corresponds to a shift operation, where the automaton consumes an input symbol from the input stream and pushes it onto the stack while transitioning to a new state. This operation reflects the fundamental step of recognizing and accepting a terminal symbol in the input, advancing the parsing process.
    \item \textbf{Reduction Edges} model reduction operations in LR(1) parsing. In traditional LR(1) parsing, reductions involve replacing a sequence of terminal symbols with a non-terminal symbol according to the grammar rules. However, nested grammar rules often require multiple reduction steps, leading to inefficiencies. Reduction edges address this by directly encoding reduction operations as single-step transitions during the pre-processing phase. These edges connect reduction targets, enabling the automaton to handle nested reductions efficiently.
    %model reduction operations, where a sequence of terminal symbols is replaced by a non-terminal. In traditional LR(1) parsing, nested grammar rules often require multiple reduction steps, introducing computational overhead. During a reduction, the automaton performs specific stack operations: it backtracks along the path of the transition, popping states from the stack until reaching the reduction endpoint. Our DPDA addresses this by introducing reduction edges, which directly encode reduction operations as single-step transitions during the pre-processing. By connecting reduction targets with these edges, the automaton can efficiently handle nested reductions, streamlining the parsing process and maintaining determinism.
\end{itemize}
% To better illustrate the distinction between acceptance edges and reduction edges, 

%In order to leverage the high-level information of the grammar, we build the DPDA of the whole grammar based on the DPDA state transition graph. The node of the DPDA is the LR(1) item set family defined in the DPDA transition graph. In order to build a DPDA, we need to consider all possible transition edges in the graph. For convenience, we categorize the edge of the DPDA into \textbf{acceptance edges} and \textbf{reduction edges}.

\subsubsection{Acceptance Edges and Reduction Edges Integration} 
The state transition graph alone cannot function as a DPDA because it only supports shift operations (\ie symbol acceptance) and lacks reduction operations, while some edges also suffer from nondeterminism. To address these issues, we not only compute all possible transition edges, including both shift and reduction edges, to complete the missing reduction paths, but also leverage prefix-conditioned edges to incorporate stack conditions into each transition, resolving nondeterminism and enabling the transformation of the non-deterministic state transition graph into a DPDA.

\textbf{Adding Acceptance Edges:} 
Acceptance edges do not need to consider determinism because the construction of the LR(1) state transition graph ensures that no node will have two identical transitions. As a result, when an acceptance edge is encountered, the target node's state information is simply pushed onto the runtime stack. The algorithmic flow of this operation is described in Lines 6–8 of Algorithm~\ref{alg:lr1_to_dpda}.

\textbf{Adding Reduction Edges:}
Based on the definition of reduction edges, we can employ a two-step method to add all necessary reduction edges to the automaton, which is described in Lines 9–18 of Algorithm~\ref{alg:lr1_to_dpda}.

% First, we identify $\epsilon$-reduction transitions, which represent unconditional reductions. These transitions are added to the automaton to handle cases where a reduction must be performed without alternative choices. During reduction, the reduction edge backtracks along the path it initially traversed, popping states from the stack until reaching the reduction endpoint. However, $\epsilon$-reduction transitions lack accept symbols, which introduces ambiguity and violates the deterministic nature of the DPDA. To ensure completeness, this process is applied recursively, generating all necessary reduction edges by traversing the state transition graph.

% Second, to resolve indeterminism, we merge these $\epsilon$-reduction edges with appropriate acceptance edges that share the same stack operations and assign suitable accept tokens. This merging process ensures that the Prefix-condition is matched, meaning the stack operations and reduction targets align correctly.
First, we identify $\epsilon$-reduction transitions, representing unconditional reductions, and add them to the automaton to handle mandatory reductions. These transitions backtrack along their path, popping states until reaching the reduction endpoint. However, their lack of accepted symbols introduces ambiguity, violating the DPDA's determinism. To ensure completeness, this process is applied recursively, generating all necessary reduction edges by traversing the state transition graph.

Second, we resolve indeterminism by merging $\epsilon$-reduction edges with compatible acceptance edges, ensuring aligned stack operations and reduction targets, and assigning appropriate accept tokens to satisfy the Prefix-condition.
%If we directly employ the previous described algorithm in the real-world LR(1) grammars, we will encounter some problems. \ref{fig:circle_issue_description} is a simple example to help us explain how the issue appear.

\subsubsection{Solving Issues with Automaton Cycles}
%LR(1) grammars are highly expressive and can handle complex language constructs, including the acceptance of cyclic symbol sequences. 
% In the automaton, this capability manifests as the presence of cycles. 
%However, new challenges arise when applying the approach above to LR(1) grammars with cycles. 
LR(1) grammars are highly expressive and can handle complex language constructs, including the acceptance of cyclic symbol sequences. However, cycles introduce significant challenges when constructing a DPDA.

\begin{figure} [t]
    \centering
    \includegraphics[width=\linewidth]{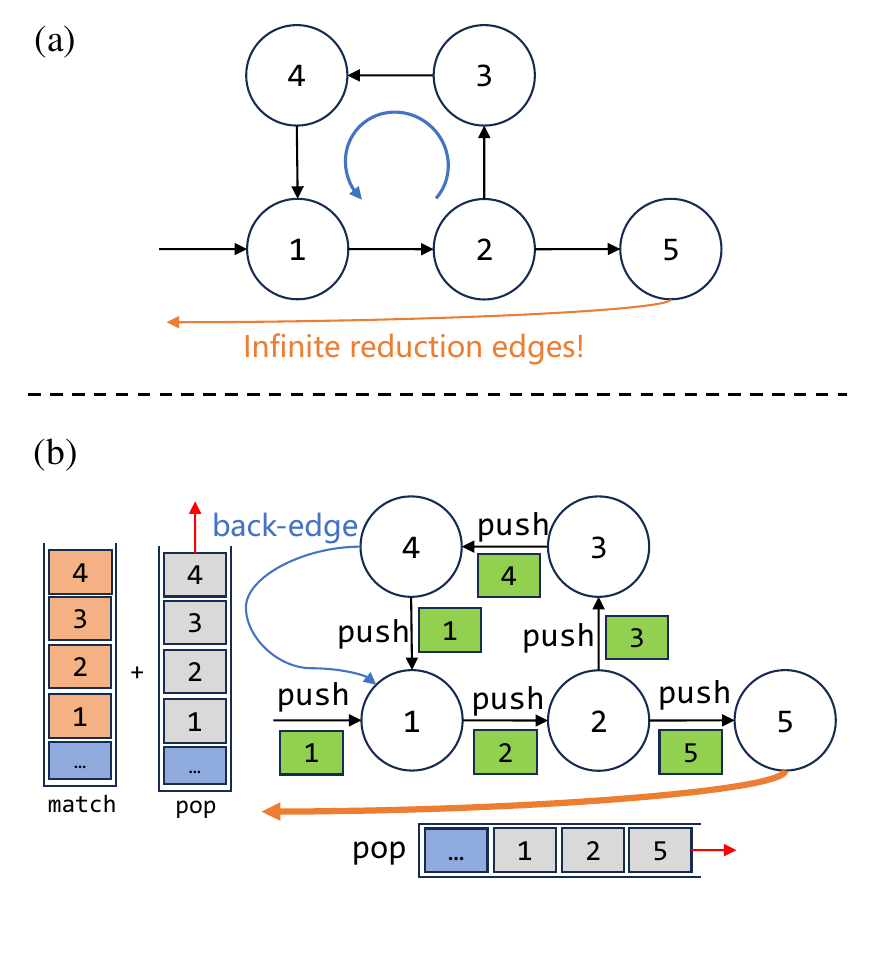}
    \caption{(a) Pushdown automaton with an infinite cycle between State 1, 2, 3, 4, leading to an infinite number of possible paths and indeterminable transition paths when adding reduction edges at State 5; (b) How our method handles the cycle issue: The back-edge from State 4 to State 1 is modified to check for complete cycle traversal information (\eg [1, 2, 3, 4]) in the stack. If detected, it pops the redundant state (\eg [1, 2, 3, 4]), ensuring reduction edges at State 5 only need to account for traversals without cycles.}
    \label{fig:circle_issue}
\end{figure}

% During the precomputation of reduction edges, cycles introduce a significant problem: Traversing a cycle repeatedly generates an infinite number of potential reduction paths, making it computationally infeasible to add all necessary reduction edges. 
% Figure~\ref{fig:circle_issue_description} visually explains how cycles in the automaton can lead to infinite reduction paths.
During the precomputation of reduction edges, cycles create a critical issue: repeatedly traversing a cycle generates an infinite number of potential reduction paths. This makes it computationally infeasible to add all necessary reduction edges. Figure~\ref{fig:circle_issue} visually illustrates how cycles in the automaton can lead to infinite reduction paths.

Through further observation, we note that during the reduction process, specifying an entry node and an exit node uniquely determines the path along which the reduction occurs. This property allows us to disregard the number of cycle traversals, as even a single traversal of the cycle does not need to be explicitly recorded.

We propose a solution that simplifies the reduction process as follows:  
Suppose we have a detected cycle with the reduction problem \( C = (s_1, s_2, s_3, \dots, s_n, s_1) \). We define the \textit{back-edge} as \( s_n \rightarrow s_1 \). While handling the cycle, we modify this back-edge by introducing an additional stack operation: a pop operation for the sequence \((s_1, s_2, \dots, s_n)\). This modification enables efficient handling of cyclic traversals.

Furthermore, by checking whether all vertices traversed in a single cycle are fully present in the execution stack, we ensure that the stack retains only the necessary information from outside the cycle traversal. Specifically, if a complete traversal of the cycle is detected, the stack information corresponding to the current traversal is popped immediately. This guarantees that the stack never accumulates redundant context from repeated cycle traversals.

%We propose a solution that simplifies the reduction process, which is shown in Figure~\ref{fig:circle_issue_solution}. Instead of tracking complete path information of the cycle within the stack, we only store the information from a single traversal of the cycle. The reduction edge is determined by identifying the entry and exit nodes of the cycle. 

%Specifically, we define the edge connecting the exit node back to the entry node as the \textit{back-edge}. During the reduction process, we add an additional transition option for the back edge. By checking whether all vertices traversed in a single cycle are fully present in the execution stack, we can determine if a complete traversal has occurred. If a match is found, the stack information from the current incomplete traversal is popped, ensuring that the stack retains only the information from a single traversal of the cycle. 
 
This approach, described in Algorithm~\ref{alg:lr1_to_dpda}, lines 1–5, guarantees that the system reverts to an equivalent state after each complete traversal, avoiding infinite reduction edges. As a result, the automaton can handle cycles efficiently without compromising determinism or computational feasibility.

\begin{algorithm} [t!]
    \SetAlgoNoEnd
    \DontPrintSemicolon
    \caption{Construct DPDA from LR(1) Transition Graph}
    \label{alg:lr1_to_dpda}

    \KwIn{LR(1) State Transition Graph $G = (S, E)$}
    \KwOut{Deterministic Pushdown Automata (DPDA)}

    \BlankLine
    \tcc{Step 1: Cycle Handling}
    $C\leftarrow$Detect cycles with reduction problem in $G$\;
    \ForEach{detected cycle $C = (s_1, s_2, ..., s_n, s_1)$}{
        \If{$C$ corresponds to recursive reduction of non-terminal $A$}{
            Define the back-edge: $s_n \xrightarrow{\text{back}} s_1$\;
            Modify the back-edge to check for complete cycle traversal in the stack: match and pop ($s_1, s_2, ..., s_n$), push($s_1$)\;
        }
    }

    \BlankLine
    \tcc{Step 2: Acceptance Edge Generation}
    \ForEach{state $s_i \in S$}{
        \ForEach{valid transition $s_i \xrightarrow{X} s_j$ in $E$}{
            Add stack operation: push($s_j$)\;
        }
    }

    % \BlankLine
    % \tcc{Step 3: Reduction Edge Generation}
    % \ForEach{state $s_i$ with reduction paths}{
    %     \tcc{Step 3.1: Identify and process unconditional reduction paths ($\epsilon$-reduction edges)}
    %     Identify all unconditional reduction paths ($\epsilon$-reduction edges) from $s_i$\;
    %     \ForEach{reduction sequence $s_i \xrightarrow{\text{reduce } A} s_j \xrightarrow{\text{reduce } B} s_k$}{
    %         Merge into a direct transition: $s_i \xrightarrow{\text{reduce } A \rightarrow B} s_k$\;
    %         Validate stack compatibility for the merged transition\;
    %     }

    %     \tcc{Step 3.2: Resolve indeterminism by merging $\epsilon$-reduction edges with acceptance edges}
    %     \ForEach{$\epsilon$-reduction edge from $s_i$}{
    %         Merge the $\epsilon$-reduction edge with appropriate acceptance edges that share the same stack operations\;
    %         Assign suitable accept tokens to ensure the Prefix-condition is matched\;
    %     }
    % }
    \tcc{Step 3: Reduction Edge Generation}
    \SetKwFunction{GenerateReductionEdges}{GenerateReductionEdges}
    \SetKwProg{Fn}{Function}{:}{}
    \Fn{\GenerateReductionEdges{state $s_i$}}{
        \ForEach{reduction sequence $s_i \xrightarrow{\text{reduce } A} s_j \xrightarrow{\text{reduce } B} s_k$}{
            Merge into a direct transition: $s_i \xrightarrow{\text{reduce } A \rightarrow B} s_k$\;
            Validate stack compatibility\;
            \GenerateReductionEdges{$s_k$}\;
        }
    
        \ForEach{$\epsilon$-reduction edge from $s_i$}{
            Merge the $\epsilon$-reduction edge with appropriate acceptance edges that share the same stack operations\;
            Assign suitable accept tokens to ensure the Prefix-condition is matched\;
            \GenerateReductionEdges{target state of the merged edge}\;
        }
    }
    \GenerateReductionEdges{initial state $s_0$}
\end{algorithm}

% Due to the presence of both terminal and non-terminal edges, the original LR(1) DPDA alone is insufficient for efficient state transitions. Handling non-terminal transitions requires scanning item sets, selecting reduction rules, backtracking along the graph, and then proceeding with the transition, introducing significant computational overhead, especially for complex grammars with large item sets and long production rules. By addressing the issue of reduction edge explosion, this optimization ensuring that the algorithm remains usable even for grammars with cyclic structures. This approach makes it well-suited for real-world applications in constrained decoding and structured text generation.
\subsection{Edge Optimization with Prefix-condition} \label{section-3-3}
\begin{figure}[t!]
    \centering
    \includegraphics[width=0.8 \linewidth]{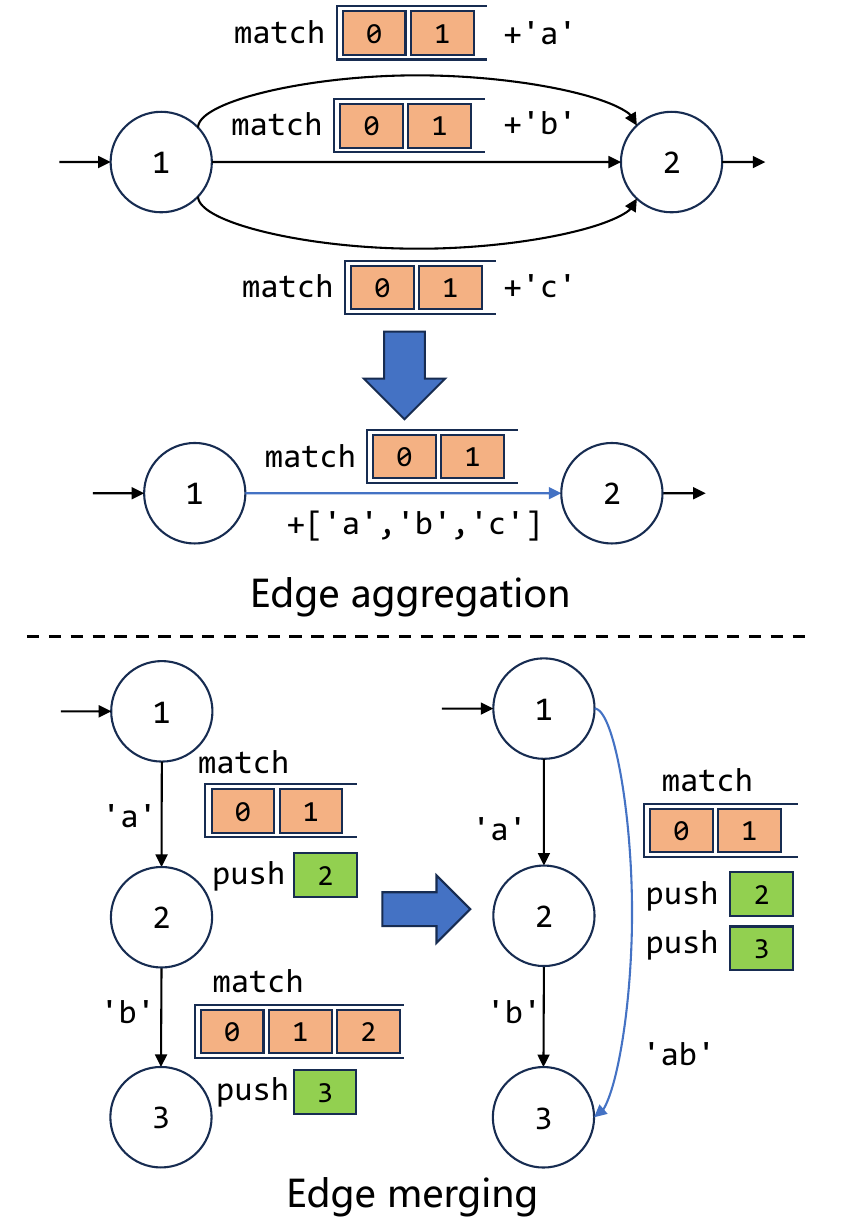}
    \caption{Two different types of edge optimization.}
    \label{fig:edge_optimization}
\end{figure}
%With Prefix-conditioned Edge Precomputation, we precompute all transitions and encode stack conditions into the edges, eliminating the need for runtime stack inspection. This not only makes the automaton fully deterministic but also enables optimizations like \textit{edge aggregation and merging} in pre-process phase, as summarized in Figure~\ref{fig:edge_optimization}.
Building on the DPDA constructed in Section~\ref{section-3-2}, we can further perform various optimizations. Since all transition edges in the DPDA are deterministic and can be uniquely resolved by matching both the stack state and input symbols, we are able to analyze the automaton's structure during the preprocessing phase. In contrast, traditional methods based on non-deterministic pushdown automata (PDA) cannot achieve such optimizations during preprocessing due to the ambiguity of transition edges, where the same input symbol may lead to multiple possible transition targets. 
For example, we can aggregate and merge transition edges as shown in Figure~\ref{fig:edge_optimization}.
\begin{itemize}[topsep=0pt,leftmargin=0.35cm]
    \item \textbf{Edge Aggregation: }Edges with the same stack prefix condition and stack operations but different accepted symbols can be combined. For example, in grammars describing numbers, edges for digits 0-9 can be merged into a single edge accepting all digits to simplify the automaton.
    \item \textbf{Edge Merging: }If two edges share the matched stack prefix condition and operations, we can connect them directly, and add a new edge. This is important for LLM constrained decoding scenarios, as it allows to ``jump'' to the desired state in fewer steps, reducing the scale of the DPDA. 
\end{itemize}

\noindent These operations are enabled by precomputed prefix-conditioned edges for all stack conditions. 
Without prefix-conditioned edges, transitions that depend on dynamic stack inspection cannot be analyzed in advance. %By combining these techniques, we further optimize the DPDA, achieving deterministic and efficient grammar parsing.

\section{Evaluation}
% In this section, we illustrate the advantages of our approach over previous work in the following areas.
% \begin{itemize}
%     \item We will evaluate our approach against previous methods at each constraint-decoding step, measuring its efficiency in structured generation tasks.
%     \item In large-batch scenarios, we will compare our method with state-of-the-art constraint decoding techniques to showcase its advantages in handling large batches.
%     \item To test practical effectiveness, we will integrate our approach into a widely used LLM inference framework, demonstrating its improvement on overall throughput.
% \end{itemize}

% \highlight{
% 1. more baselines (e.g. outlines, other baselines in XGrammar paper) \\
% 2. more model results (different type, different size?) \\
% 3. more hardware setup (A100, H100, A800, H800?) \\
% 4. end-to-end serving throughput?
% 5. different grammar in a batch, different prefill length/decode length in a batch?
% }
\subsection{Experimental Setup}
\paragraph{Implementation:}
We implemented our approach in 2,000 lines of Python code and about 1,000 lines of C++ code, and we seamlessly integrated with LightLLM~\cite{lightllm}, a popular LLM inference framework. 

%\footnotetext[1]{\url{https://github.com/ModelTC/lightllm}}

\paragraph{Hardware Setup:}
All the experiments are tested on a server with Intel(R) Xeon(R) Gold 6448Y CPU and 8 NVIDIA H800 GPUs. Depending on the scale of the experiment, we use different numbers of GPUs.

\paragraph{Baselines:} We choose the following representative works on grammar constraint decoding.
\begin{itemize}[noitemsep,topsep=1pt,leftmargin=*]
    \item \textbf{XGrammar:} An open-source library for structured generation in large-language models. It significantly enhances performance in tasks like JSON grammar generation with reduced latency and storage. 
    \item \textbf{Outlines:} A text generation library, it offers a Python tool for grammar-guided generation, offering a fast generation method. We use vLLM integrated with Outlines for evaluation.
    \item \textbf{Llama.cpp}: A C/C++-based LLM inference tool, and also includes support for grammar constraint decoding.
\end{itemize}

\paragraph{Datasets:}  
In our experiments, we utilized the JSON-mode-eval~\cite{nous2024json} dataset from NousResearch as prompts. As there is a scarcity of datasets for structured output, we collected some private data additionally and incorporated it into the test dataset.

\subsection{Per-step Decoding Efficiency}
\begin{figure}[t!]
    \centering
    \includegraphics[width=1\linewidth]{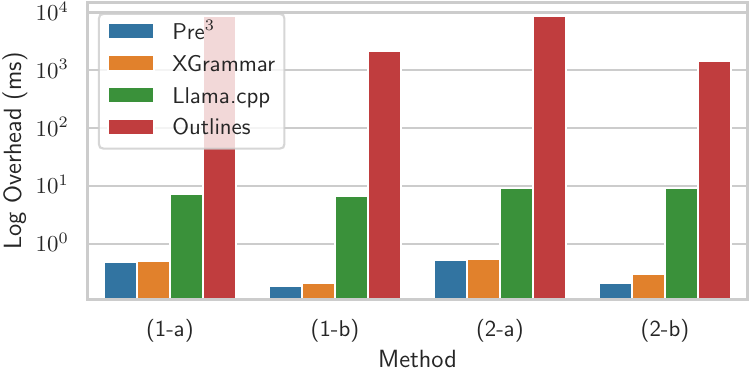}
    \caption{Per-step decoding overhead across different grammars and models. Outlines incurs an overhead of up to several seconds per step. Experiments contain (1) Chain-of-Thought grammar, (2) JSON grammar. Models contain (a) Meta-Llama-3-8B on 1$\times$H800, (b) Meta-Llama-2-70B on 4$\times$H800.}
    \label{fig:per_step_decoding_res}
\end{figure}
To evaluate the improvement of our system, we first examine the per-step decoding overhead, which is defined by subtracting the original decoding time from the grammar-based decoding time. We design four experiment setups, including two models, Meta-Llama-3-8B and Meta-Llama-2-70B, and two grammars with English characters, JSON and chain-of-thought. For comparison, we benchmark our method against several state-of-the-art and popular structure generation engines, including XGrammar, Outlines, and llama.cpp, to demonstrate the efficiency of our system at a per-step scale.

The results are shown in Figure~\ref{fig:per_step_decoding_res} and Table~\ref{tb:per-step}. \method demonstrates a superiority over Outlines and llama.cpp, and \method remains a consistent advantage over XGrammar. The results indicate that \method introduces less overhead than previous SOTA systems.

\begin{table*}[t!]
\footnotesize
\caption{Decode batch inference time comparison between our method and XGrammar. The ``-'' marker stands for the batch size cannot be executed on the given hardware setup.}
\label{tb:large-batch}
\centering
\resizebox{0.95\linewidth}{!}{
\begin{tabular}{@{}c|cccccccc@{}}
\toprule
Evaluation Configuration & Method                              & 16    & 32    & 64    & 128   & 256   & 512   & 1024  \\ \midrule
\multirow{2}{*}{Llama-3-8B~\cite{dubey2024llama}}             & XGrammar (ms)  & 15.19 & 43.69 & 52.07 & 65.21 & 90.98 & 147.64 & 272.77 \\
                                             & \method (ms)  & 11.77 & 31.12 & 35.88 & 45.32 & 64.42 & 104.46 & 201.16 \\
2$\times$H800                                & Reduction & \textbf{↓22.49\%} & \textbf{↓28.78\%} & \textbf{↓30.09\%} & \textbf{↓30.50\%} & \textbf{↓29.20\%} & \textbf{↓29.24\%} & \textbf{↓26.25\%} \\ \midrule
\multirow{2}{*}{DeepSeek-V2-Lite-Chat~\cite{liu2024deepseek}} & XGrammar (ms)  & 51.76 & 59.45 & 77.74 & 104.06 & 121.46 & -   & -   \\
                                             & \method (ms)  & 49.91 & 53.71 & 54.41 & 61.63 & 75.47 & -   & -   \\
15.7B\quad 2$\times$H800                          & Reduction & \textbf{↓3.57\%}  & \textbf{↓9.65\%}  & \textbf{↓30.01\%} & \textbf{↓40.78\%} & \textbf{↓37.86\%} & -   & -   \\ \midrule
\multirow{2}{*}{Qwen2-14B~\cite{yang2024qwen2}}                  & XGrammar (ms)  & 16.77 & 47.94 & 57.05 & 74.54 & 98.64 & 162.47 & 285.42 \\
                                             & \method (ms)  & 16.52 & 47.94 & 47.89 & 65.50 & 90.20 & 143.83 & 232.18 \\
INT8\quad 2$\times$H800                               & Reduction & \textbf{↓1.52\%} & \textbf{↓0.12\%} & \textbf{↓2.37\%}  & \textbf{↓12.14\%} & \textbf{↓8.55\%}  & \textbf{↓11.47\%} & \textbf{↓18.65\%} \\ \midrule
\multirow{2}{*}{Llama-2-70B~\cite{touvron2023llama}}            & XGrammar (ms)  & 28.75 & 55.12 & 56.94 & 68.79 & 85.92 & -   & -   \\
                                             & \method (ms)  & 27.20 & 54.24 & 54.18 & 62.27 & 75.72 & -   & -   \\
4$\times$H800                          & Reduction & \textbf{↓5.39\%}  & \textbf{↓1.60\%}  & \textbf{↓4.85\%}  & \textbf{↓9.48\%}  & \textbf{↓11.87\%} & -   & -   \\ \bottomrule
\end{tabular}
}
\end{table*}

\begin{table}[h]
\centering
\caption{Per-step decode time comparison between our method and XGrammar.}
\label{tb:per-step}
\resizebox{\linewidth}{!}{
\begin{tabular}{ccc|cc}
\toprule
 & \multicolumn{2}{c}{Llama-3-8B} & \multicolumn{2}{c}{Llama-2-70B} \\
\midrule
Batchsize & \method & XGrammar & \method & XGrammar \\
1 & \textbf{0.5172} & 0.5531 & \textbf{0.2163} & 0.3030 \\
4 & \textbf{0.6537} & 0.9327 & \textbf{0.2407} & 0.3310 \\
\bottomrule
\end{tabular}
}
\label{tab:per-step-comparer} % Optional label for referencing the table
\end{table}

% Specifically, we measure the time taken for grammar decoding and compare it to the time required for standard decoding, with the goal of demonstrating that our method offers a practical improvement in terms of efficiency. 这段跟前面有些重复

\subsection{Large Batch Inference Efficiency}
\begin{figure*}[tp!]
    \centering
    \includegraphics[width=0.95\linewidth]{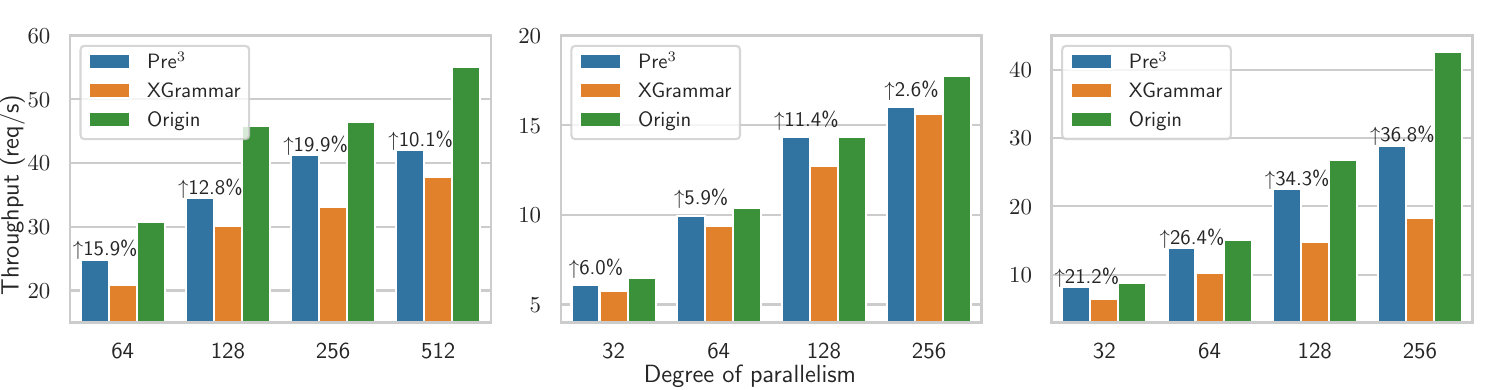}
    \caption{System throughput based on different models and concurrency levels. Left: Llama3-8B, Middle: Llama2-70B, Right: DeepSeek-V2-Lite-Chat.}
    \label{fig:throughput}
\end{figure*}

% In real-world serving scenarios, inference often involves handling large batches of requests simultaneously, making large-batch efficiency a crucial factor for deploying language models at scale in production. In this context, we focus on evaluating the performance in large-batch settings, where efficiency gains can have a significant impact on overall system performance.

% We benchmarked the efficiency of our approach \method against the current state-of-the-art, XGrammar. For the workload, we chose the JSON grammar because it is more complex and challenging for efficient grammar engines. The context-free grammar of JSON is more intricate with recursive structures such as lists and dictionaries, while the JSON schema is more constrained and less recursive. 
In real-world serving scenarios, inference often handles large batches of requests simultaneously, making large-batch efficiency crucial for deploying language models at scale. We evaluate performance in such settings, where efficiency gains significantly impact system performance.

We benchmark \method against the state-of-the-art XGrammar, using the JSON grammar for its complexity and challenging recursive structures (\eg lists and dictionaries). This tests the robustness and scalability of our method under demanding conditions.
% Given this, the JSON grammar provides a more stringent test of the efficiency of grammar-based decoders.

Our experiments are conducted on multiple models of varying sizes and architectures. Specifically, we conducted experiments on Llama3-8B and Deepseek-V2 (15.7B) on a 2$\times$H800 setup, and Llama2-70B on a 4$\times$H800 setup. The maximum batch size goes to 1024, large enough to test the scalability of our method.
In this experiment, we also measured the average time taken for each step, but the requests are batched in number to test the system's ability to process large batches.

The result is shown in Table~\ref{tb:large-batch}. The results show that \method consistently outperforms XGrammar in all scenarios with latency reduction by up to 30\%. The advantage is more significant at larger batch sizes, demonstrating the scalability of \method.

\subsection{Realworld Deployment}
To evaluate the throughput in real-world service environments, we compare the performance of XGrammar and \method, under varying system concurrency levels. We conducted simulation experiments on Meta-Llama-3-8B (2$\times$H800) and Meta-Llama-2-70B (4$\times$H800), measuring the throughput in burst scenarios at different levels of concurrency.

The results are shown in Figure~\ref{fig:throughput}. Both \method and XGrammar have lower throughput than the Original system due to the added overhead introduced by constraint decoding, while \method demonstrated a significant improvement over XGrammar, achieving up to 20\% higher throughput at higher concurrency levels, showing that \method provides higher throughput in end-to-end deployment.
\section{Related Work}
\paragraph{LLM Constrained Decoding.}
Several approaches have been proposed for constrained decoding in language models, yet most exhibit limitations when applied to large-batch inference. 
Implementations such as llama.cpp~\cite{llama_cpp} rely on inefficient runtime token verification, introducing significant computational overhead. Subsequent methods like Outlines~\cite{willard2023efficientguidedgenerationlarge} and SynCode~\cite{ugare2024syncodellmgenerationgrammar} improve upon grammar-guided generation but still face suboptimal decoding efficiency. The current state-of-the-art method, XGrammar~\cite{dong2024xgrammarflexibleefficientstructured}, achieves impressive speed for small batch sizes; however, its performance degrades as batch sizes increase due to growing computational overhead. Similarly, GreatGrammar~\cite{park2025flexibleefficientgrammarconstraineddecoding} demonstrates strong efficiency in handling complex grammars but is only evaluated with batch size equals to 1, leaving its scalability to larger batches an open question.

\paragraph{LR(1) Grammar Parser.}
The theoretical foundations of LR(1) parsing and its equivalence to deterministic pushdown automata (DPDA) have been well-established in formal language theory and compiler design~\cite{deremer1969practical, lehmann1971lr, korenjak1969practical}. Traditional LR(1) parsers, such as those described by Knuth~\cite{knuth1965translation}, use state-merging techniques to construct minimal parsing tables, enabling deterministic recognition of context-free languages. 
Recent work, including IELR(1)~\cite{denny2008ielr} and PSLR(1)~\cite{denny2010pslr}, further optimized the parser by addressing state conflicts and improving efficiency in handling composite grammars. These methods ensure that LR(1) parsers can be systematically converted into DPDA implementations, where a deterministic state transition table guides stack operations.
While prior work has focused on compiler parsing, applying LR(1)-to-DPDA techniques to constrained decoding in large language models (LLMs) poses unique challenges. To our knowledge, \method presents the first adaptation of LR(1) parsing techniques to the domain of LLM constrained decoding. 
\section{Conclusion}
% In this work, we recognized the overlooked structural grammar properties in structured generation services. To exploit this, we presented \method, a system that accelerates constrained decoding by precomputing and optimizing unified automation constructed from grammar. \method significantly outperforms existing SOTA baselines by up to 30\% in throughput and demonstrates greater advantages with large batch sizes. 

This work addressed the limitations of existing structured generation approaches by proposing a DPDA-based methodology (\method), which integrates Cycle-aware Deterministic Pushdown Automata Construction and Prefix-conditioned Edge Optimization. 
\method significantly outperforms existing SOTA baselines by up to 40\% in throughput and exhibits greater advantages with large batch sizes.

\section*{Limitation}

While our work demonstrates significant improvements in constrained LLM decoding efficiency, several limitations and potential areas for improvement remain.
Firstly, our method is optimized for LR(1) grammars, which cover most structured generation needs, but faces challenges with more complex LR($k$) grammars ($k > 1$). These require intricate state transitions and lookahead mechanisms, increasing DPDA construction and processing complexity. Future work should explore hybrid parsing or adaptive mechanisms to handle such cases efficiently.
Second, our Python-based research prototype lacks production-level optimizations. Although suitable for experimentation, the implementation could benefit from hardware acceleration (\eg GPU parallelization for grammar processing) and system-level optimizations. A future C++/Rust implementation with fine-tuned memory management could significantly improve performance and scalability. Although preprocessing complexity scales with grammar size (due to increased edges and cycles), current processing times remain practical for real-world deployment.
Addressing these limitations could unlock additional performance improvements and broaden the applicability of our approach.
\section*{Acknowledgements}
This work was partially sponsored by the National Key R\&D Program of China (No. 2022ZD0119100), in part by China NSF grant No. 62472278, 62025204, 62432007, 62441236, 62332014, 62332013, and 62322201, in part by Alibaba Group through Alibaba Innovation Research Program, and in part by Tencent Rhino Bird Key Research Project. This work was partially supported by SJTU Kunpeng \& Ascend Center of Excellence.
The opinions, findings, conclusions, and recommendations expressed in this paper are those of the authors and do not necessarily reflect the views of the funding agencies or the government.

% \highlight{
% long paper structure: \\
% abstract 1/2 column \\
% introduction 1.5 pages \\
% method 2.5 pages (graph: overview + 2-3 detail) \\
% experiment 3 pages (with graph and table) \\
% conclusion 1/2 column \\
% related work 2/3 column \\
% limitation 1/2 column \\
% }

% \section*{Acknowledgments}
% This document has been adapted
% by Steven Bethard, Ryan Cotterell and Rui Yan
% from the instructions for earlier ACL and NAACL proceedings, including those for
% ACL 2019 by Douwe Kiela and Ivan Vuli\'{c},
% NAACL 2019 by Stephanie Lukin and Alla Roskovskaya,
% ACL 2018 by Shay Cohen, Kevin Gimpel, and Wei Lu,
% NAACL 2018 by Margaret Mitchell and Stephanie Lukin,
% Bib\TeX{} suggestions for (NA)ACL 2017/2018 from Jason Eisner,
% ACL 2017 by Dan Gildea and Min-Yen Kan,
% NAACL 2017 by Margaret Mitchell,
% ACL 2012 by Maggie Li and Michael White,
% ACL 2010 by Jing-Shin Chang and Philipp Koehn,
% ACL 2008 by Johanna D. Moore, Simone Teufel, James Allan, and Sadaoki Furui,
% ACL 2005 by Hwee Tou Ng and Kemal Oflazer,
% ACL 2002 by Eugene Charniak and Dekang Lin,
% and earlier ACL and EACL formats written by several people, including
% John Chen, Henry S. Thompson and Donald Walker.
% Additional elements were taken from the formatting instructions of the \emph{International Joint Conference on Artificial Intelligence} and the \emph{Conference on Computer Vision and Pattern Recognition}.


\newpage
\bibliography{custom}

\newpage
\appendix
\onecolumn
\section{Supplementary Materials on Formal Language Theory}\label{sec:appendix-1}
To facilitate understanding of our algorithm, we introduce some basics of formal language theory in the supplementary materials.

\subsection{Formal Definition of LR(1) Grammars}

LR(1) grammars constitute a fundamental class of context-free grammars that can be parsed deterministically with one-symbol lookahead. Formally, a grammar \( G = (V, \Sigma, P, S) \) is said to be LR(1) if the following conditions are satisfied:
\begin{itemize}
    \item \textbf{Uniqueness of Valid Items:} For any viable prefix \(\gamma\), there exists at most one LR(1) item of the form \([A \rightarrow \alpha \cdot \beta, a]\), where \( a \in \Sigma \cup \{\$\} \), that is valid at that parsing configuration.
    \item \textbf{Reduction Consistency:} If two items \([A \rightarrow \alpha \cdot, a]\) and \([B \rightarrow \beta \cdot, b]\) are simultaneously valid for the same viable prefix \(\gamma\), then at least one of the following must hold: \( A \neq B \), \( \alpha \neq \beta \), or \( a \neq b \).
\end{itemize}
These constraints ensure that shift/reduce and reduce/reduce conflicts are resolved uniquely using the lookahead token. All LR(0) grammars are properly contained within the LR(1) class, and the parsing process maintains linear time complexity with respect to the input length.

\subsection{Detailed Construction Procedure of the Canonical LR(1) Automaton}

The construction of the canonical LR(1) automaton is a central step in generating a deterministic parser for a given context-free grammar. The automaton is a finite state machine in which each state represents a set of LR(1) items, and transitions correspond to the recognition of grammar symbols.

Let \( G = (V, \Sigma, P, S) \) be the original grammar. We begin by augmenting it with a new start symbol \( S' \notin V \), and add the production \( S' \rightarrow S \). The augmented grammar is denoted by \( G' = (V \cup \{S'\}, \Sigma, P \cup \{S' \rightarrow S\}, S') \).

\paragraph{LR(1) Item}

An LR(1) item is a pair \([A \rightarrow \alpha \cdot \beta, a]\), where:
\begin{itemize}
    \item \( A \rightarrow \alpha \beta \) is a production rule in \( P \),
    \item The dot (\(\cdot\)) indicates the current parsing position within the right-hand side,
    \item \( a \in \Sigma \cup \{\$\} \) is a lookahead symbol, representing the terminal expected to follow the derivation of \( A \).
\end{itemize}

\paragraph{Item Set (State)}

A state in the automaton is a set of LR(1) items, closed under the \texttt{CLOSURE} operation. Each state encapsulates a snapshot of all valid parser configurations at a certain point in the input.

\paragraph{Closure Operation}

The \texttt{CLOSURE} function expands a set of items \( I \) by recursively adding all items that could be expected next due to nonterminal symbols appearing after the dot. Formally:
\[
\texttt{CLOSURE}(I) = \text{smallest superset of } I
\]
such that:
\[
\forall [A \rightarrow \alpha \cdot B \beta, a] \in I,\, \forall B \rightarrow \gamma \in P,\, \forall b \in \text{FIRST}(\beta a):\ [B \rightarrow \cdot \gamma, b] \in \texttt{CLOSURE}(I)
\]

The key detail here is that the lookahead symbol \( a \) propagates through the \(\text{FIRST}(\beta a)\) computation, ensuring context-sensitivity.

\paragraph{Goto Operation}

Given a state \( I \) and a grammar symbol \( X \in V \cup \Sigma \), the \texttt{GOTO} function computes the next state by advancing the dot over \( X \) in all applicable items:
\[
\texttt{GOTO}(I, X) = \texttt{CLOSURE}\left(\left\{[A \rightarrow \alpha X \cdot \beta, a] \mid [A \rightarrow \alpha \cdot X \beta, a] \in I\right\}\right)
\]

This represents the parser consuming symbol \( X \) and transitioning to a new parser configuration.

\paragraph{Construction Algorithm}

Let \( C \) denotes the set of item sets (states). The construction algorithm proceeds as follows:

\begin{enumerate}
  \item Initialize:
  \[
  I_0 = \texttt{CLOSURE}(\{[S' \rightarrow \cdot S, \$]\}),\quad C := \{I_0\}
  \]
  
  \item Iteratively expand:
  
  For each \( I \in C \), and each grammar symbol \( X \in V \cup \Sigma \):
  \[
  J := \texttt{GOTO}(I, X)
  \]
  If \( J \neq \emptyset \) and \( J \notin C \), then add \( J \) to \( C \) and record a transition:
  \[
  \delta(I, X) = J
  \]
  
  \item Repeat until no new item sets are added to \( C \).
\end{enumerate}

\paragraph{Parsing with the Canonical LR(1) Automaton}

Once the canonical LR(1) automaton and corresponding parsing tables have been constructed, the LR(1) parser performs deterministic syntax analysis by simulating a left-to-right scan of the input, using a stack-based mechanism.

At the core of the parsing process are two tables derived from the automaton:

\textbf{ACTION} table maps each parser state and terminal symbol (including the end-of-input symbol \$) to one of four possible actions:
\begin{itemize}
  \item \emph{Shift \(s_j\)}: Advance to state \(j\) by consuming the current input symbol.
  \item \emph{Reduce \(r_k\)}: Apply production rule \( A \rightarrow \beta \) and reduce the right-hand side.
  \item \emph{Accept}: Recognize the input as belonging to the language, which occurs only when the parser encounters the item \( [S' \rightarrow S\cdot, \$] \).
  \item \emph{Error}: No valid action exists for the current state and input; a syntax error is reported.
\end{itemize}

\textbf{GOTO} table maps each state and non-terminal symbol to a successor state, and is only used after reductions to determine the next parser state following a non-terminal transition.

\paragraph{Stack-based Parsing Mechanism}

The parser maintains a stack \( \mathcal{S} \), initially containing only the start state \( s_0 \). The input string \( w = a_1a_2\ldots a_n \) is augmented with the end-of-input marker \( \$ \), and a pointer is maintained to the current input symbol.

Each entry in the stack alternates between a grammar symbol and a state, \ie
\[
\mathcal{S} = [X_0, s_0, X_1, s_1, \dots, X_k, s_k]
\]

The parsing loop proceeds as follows:

\begin{enumerate}
  \item Let \( s_k \) be the state at the top of the stack, and let \( a \) be the current input symbol.
  
  \item Consult the ACTION table at entry \( \text{ACTION}[s_k, a] \):
  \begin{itemize}
    \item \emph{If it specifies Shift \( s_j \)}: Push the input symbol \( a \) and state \( s_j \) onto the stack, and advance the input pointer.
    
    \item \emph{If it specifies Reduce by production \( A \rightarrow \beta \)}: Pop \( 2|\beta| \) entries from the stack (removing both symbols and states), exposing the new top state \( s' \). Push \( A \), and then consult \( \text{GOTO}[s', A] = s'' \) to push the new state \( s'' \).
    
    \item \emph{If it specifies Accept}: The parser halts and returns success, confirming that the input belongs to the language defined by the grammar.
    
    \item \emph{If no action is defined (Error)}: The parser halts and reports a syntax error.
  \end{itemize}
  
  \item Repeat the above steps until either an Accept or Error action is encountered.
\end{enumerate}

\subsection{Definition of Pushdown Automata and Deterministic Pushdown Automata}

A \emph{pushdown automaton} (PDA) is a computational model that extends a finite automaton with a stack memory.  Formally, a (nondeterministic) PDA is defined as a 7-tuple
$M = (Q, \Sigma, \Gamma, \delta, q_0, Z_0, F)$,
where:
\begin{itemize}
\item $Q$ is a finite set of states,
\item $\Sigma$ is a finite input alphabet,
\item $\Gamma$ is a finite stack alphabet,
\item $\delta: Q \times (\Sigma \cup {\epsilon}) \times \Gamma ;\to; \mathcal{P}(Q \times \Gamma^)$ is the transition function,
\item $q_0 \in Q$ is the start state,
\item $Z_0 \in \Gamma$ is the initial stack symbol,
\item $F \subseteq Q$ is the set of accepting (final) states.
\end{itemize}
Here $\epsilon$ denotes the empty string and $\mathcal{P}(X)$ is the powerset of $X$.  Intuitively, the PDA reads one symbol at a time (or makes $\epsilon$-moves without consuming input), and at each step it may push or pop symbols on the stack. A transition of the form
$\delta(q, a, A) \ni (p, \gamma)$
means that if the automaton is in state $q$, sees input symbol $a$ (or $a=\epsilon$ for an $\epsilon$-move), and $A$ is the top stack symbol, then it can move to state $p$, pop $A$, and push the string $\gamma\in\Gamma$ onto the stack (with the leftmost character of $\gamma$ becoming the new top).  A string is accepted if the PDA can consume the entire input and reach a configuration in which the current state is in $F$ (accepting), regardless of the remaining stack content.  (Alternatively, acceptance by empty stack can be used; for nondeterministic PDAs these two acceptance modes yield the same class of languages.)

A \emph{deterministic pushdown automaton} (DPDA) is a special kind of PDA with restrictions that eliminate nondeterminism.  Formally, a DPDA is defined by the same 7-tuple structure, but its transition function $\delta$ must satisfy determinism conditions: for any state $q \in Q$ and stack symbol $A \in \Gamma$, at most one of the following can occur:
\begin{itemize}
\item There is at most one input symbol $a \in \Sigma$ such that $\delta(q,a,A)$ is nonempty (so for each fixed $(q,A)$, there cannot be two different input symbols leading to transitions).
\item If $\delta(q,\epsilon,A)$ is nonempty (an $\epsilon$-move is available), then $\delta(q,a,A)$ must be empty for every $a\in\Sigma$ (so that no input-consuming move competes with an $\epsilon$-move on the same $(q,A)$).
\end{itemize}
Informally, in a DPDA, the next move is uniquely determined by the current state, the current input symbol (or $\epsilon$), and the top of the stack.  Equivalently, for each $(q,A)$ there is at most one transition available in total, and it cannot be both an $\epsilon$-move and a non-$\epsilon$-move simultaneously.  A DPDA typically accepts by reaching an accepting state in $F$ after consuming all input.  A language is called \emph{deterministic context-free} (a DCFL) if it is recognized by some DPDA; otherwise it may require a nondeterministic PDA.

\subsection{Language Classes: CFL vs DCFL}

The class of languages recognized by PDAs (nondeterministic) is exactly the class of \emph{context-free languages} (CFLs).  By contrast, the class of languages recognized by DPDAs is the class of \emph{deterministic context-free languages} (DCFLs).  It is known that DCFLs are a strict subset of CFLs: there are context-free languages that no DPDA can recognize.  In general, every DPDA is also a PDA (so DCFL~$\subseteq$~CFL), but many CFLs require nondeterminism.

Some key differences and properties include:
\begin{itemize}
\item \textbf{Expressive power:}  Every DCFL is context-free, but there are CFLs that are not deterministic.  For example, the language of all palindromes ${ww^R : w \in {a,b}^*}$ is context-free but not deterministic context-free (no DPDA can decide the midpoint of the string to switch from pushing to popping).
\item \textbf{Unambiguity:} Every DCFL has an unambiguous context-free grammar and a unique leftmost derivation for each string.  In contrast, CFLs in general may be ambiguous or inherently nondeterministic.  (In fact, if a context-free language has a deterministic PDA, it admits no ambiguity in parsing.)
\item \textbf{Closure properties:} DCFLs enjoy stronger closure than general CFLs.  For instance, DCFLs are closed under complementation (by converting final-state acceptance to empty-stack acceptance and flipping accepting conditions), whereas CFLs are not closed under complement in general.  Also, DCFLs are closed under intersection with regular languages.  In contrast, CFLs are not closed under complement or intersection in general.
\item \textbf{Parsing and complexity:} Deterministic PDAs can be executed in linear time, and they correspond to deterministic parsing algorithms (such as LL(1) or LR(1) parsers for programming languages).  Nondeterministic PDAs also run in linear time (in theory) but require nondeterminism or backtracking to decide moves.
\end{itemize}

\subsection{Transition Function and Determinism Conditions}

The transition function $\delta$ of a PDA (resp.\ DPDA) encodes its moves.  Recall
$\delta: Q \times (\Sigma \cup \{\epsilon\}) \times \Gamma \;\to\; \mathcal{P}(Q \times \Gamma^*)$.
If $(p,\gamma)\in \delta(q,a,A)$, this means the PDA can move from configuration $(q,aw,\dots)$ to $(p,w,\dots)$ by consuming $a$ (unless $a=\epsilon$) and replacing the stack top $A$ with the string $\gamma$.  Concretely, if the current configuration is $(q,av,\alpha A)$ (state $q$, remaining input $av$, stack $\alpha A$ with $A$ at top), then after the transition it goes to $(p,v,\alpha\gamma)$ (state $p$, remaining input $v$, stack $\alpha\gamma$).  The string $\gamma$ may be empty (denoted $\epsilon$), which corresponds to simply popping $A$ without pushing anything.

For a nondeterministic PDA, there may be multiple choices $(p,\gamma)$ in the set $\delta(q,a,A)$, reflecting different possible moves.  A DPDA imposes the restriction that these choices must be essentially unique.  In particular:
\begin{itemize}
\item For each state $q$ and stack symbol $A$, and for each input symbol $a\in\Sigma$, there is at most one pair $(p,\gamma)\in \delta(q,a,A)$.  That is, the PDA cannot have two different transitions that read the same input $a$ in the same state $q$ with the same stack top $A$.
\item Moreover, if $\delta(q,\epsilon,A)$ is nonempty (\ie an $\epsilon$-move is available from $(q,A)$), then $\delta(q,a,A)$ must be empty for every $a\in\Sigma$.  Equivalently, from any configuration $(q,A)$, the machine cannot both use an $\epsilon$-move and a non-$\epsilon$-move; at most one type of move is allowed.
\end{itemize}
These conditions ensure that at most one transition is available in any situation, making the automaton deterministic.  In practice, a DPDA’s transition function is often given as a function rather than a relation, since there is at most one output move for each $(q,a,A)$ (or $(q,\epsilon,A)$).

\end{document}